\documentclass[letterpaper, 10 pt, conference]{ieeeconf}  % Comment this line out if you need a4paper

\IEEEoverridecommandlockouts                              % This command is only needed if 
\usepackage{graphicx} % for pdf, bitmapped graphics files
\usepackage{amsmath} % assumes amsmath package installed
\usepackage{amssymb}  % assumes amsmath package installed
\usepackage{acronym}
\usepackage{xcolor}
\usepackage{textcomp}
\usepackage{xspace}
\usepackage{multirow}      % per \multirow
\usepackage{graphicx}      % per \resizebox e \rotatebox
\usepackage{booktabs}     
\usepackage{hyperref}
\usepackage{booktabs}
\usepackage{multirow}
\usepackage{comment}
\usepackage{capt-of}
\hypersetup{
    colorlinks=true,
    linkcolor=blue,
    filecolor=magenta,      
    urlcolor=cyan,
    pdftitle={Overleaf Example},
    pdfpagemode=FullScreen,
    }
\newcommand{\eg}{\emph{e.g.,}\xspace}

\newcommand{\boldparagraph}[1]{\textbf{#1.}}
\newcommand{\rrf}{\emph{R2F}}
\newcommand{\rrfv}{\emph{R2F-VLN}}

\acrodef{LLM}{Large Language Model}
\acrodef{VLM}{Visual Language Model}
\acrodef{VLN}{Visual-Language Navigation}
\acrodef{ObjectNav}{Object-Goal Navigation}
\acrodef{SVM}{Semantic Voxel Map}
\acrodef{SRF}{Semantic Ray Frontier}
\acrodef{ViT}{Vision Transformer}
\acrodef{OOR}{out-of-range}
\acrodef{NLP}{Natural Language Processing}
\acrodef{FM}{Foundation Model}

\acrodef{SR}{Success Rate}
\acrodef{SPL}{Success weighted by Path Length}

\title{\LARGE \bf
R2F: Repurposing Ray Frontiers for LLM-free Open-Vocabulary Object Navigation 
}

\author{Francesco Argenziano$^{1}$, John M. A. Marcelo$^{1}$, Michele Brienza$^{1}$, Abdel Hakim Drid$^{2}$, Emanuele Musumeci$^{1}$,\\ Domenico D. Bloisi$^{3}$, Daniele Nardi$^{1}$, and Vincenzo Suriani$^{1}$ % 
\thanks{$^{1}$Department of Computer, Automation and Management Engineering,
        Sapienza University of Rome, Rome, Italy
        {\tt\small lastname@diag.uniroma1.it},
        {\tt\small  marcelo.1850518@studenti.uniroma1.it}, 
        $^{2}$Department of Electronics and Automation, Mohamed Khider University of Biskra, Biskra, Algeria
        {\tt\small abdelhakim.drid@univ-biskra.dz}
        $^{3}$International University of Rome UNINT, Rome, Italy
        {\tt\small domenico.bloisi@unint.eu}}%
}

\begin{document}

\makeatletter
\let\@oldmaketitle\@maketitle
\renewcommand{\@maketitle}{\@oldmaketitle
\centering
% \vspace{-0.5em}
% \vspace{1em}
\includegraphics[width=0.85\linewidth]{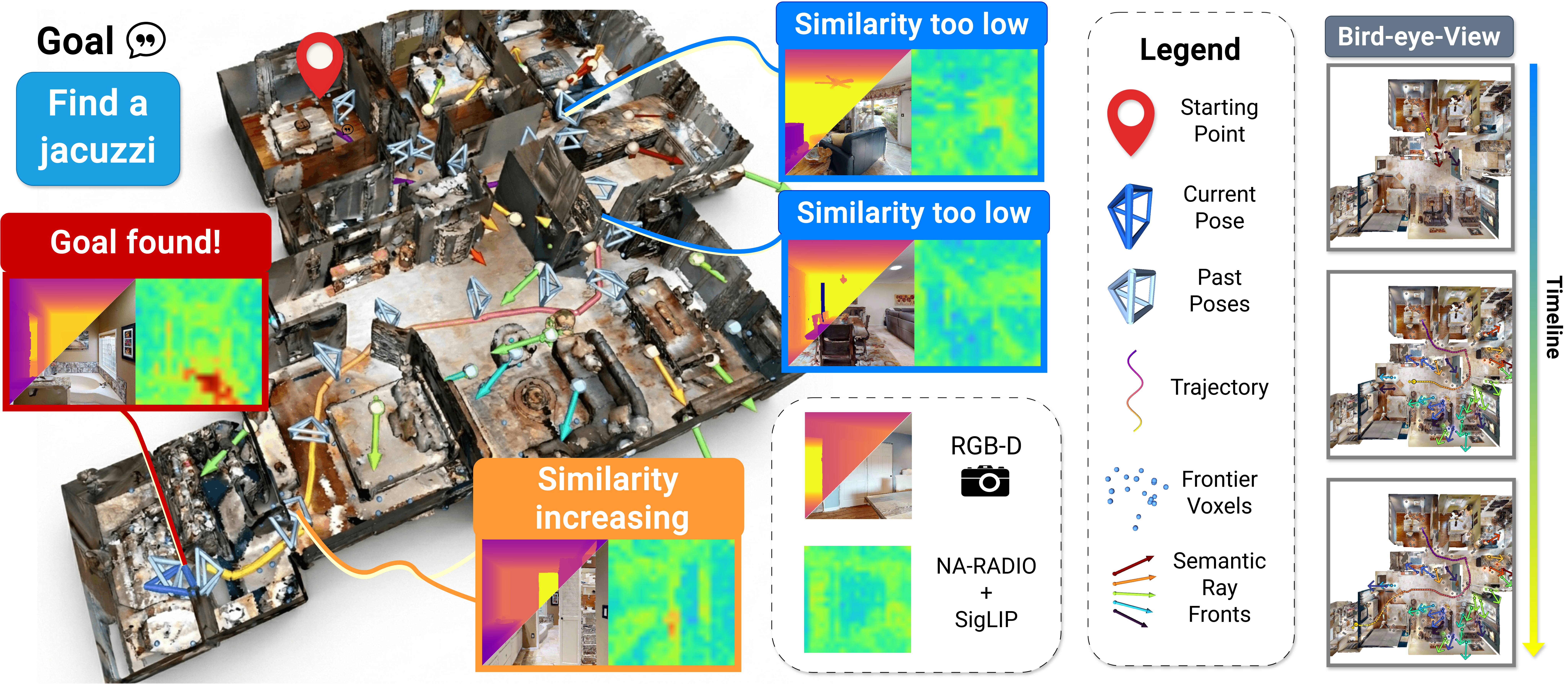}
\setcounter{figure}{0} % <-- Reset figure counter
\refstepcounter{figure} % Ensures the counter advances properly
\addtocounter{figure}{-1} % If needed, adjust the numbering
\captionof{figure}{\textbf{\rrf} at a glance. A representative zero-shot open-vocabulary object navigation episode in a photorealistic indoor scene. As the agent explores, open-vocabulary semantic evidence is accumulated along out-of-range rays and attached to frontier regions (visualized as cosine-similarity heatmaps). Frontier scores increase in directions consistent with the target query, enabling embedding-based subgoal selection without iterative LLM/VLM deliberation. The episode terminates when the target is confidently detected (\emph{Goal found!}). Code and supplementary material are available at \url{https://lab-rococo-sapienza.github.io/r2f/}.}

\label{fig:splash}
\vspace{-0.98em}
}
\makeatother

% % Explicitly reset figure counter after title
\setcounter{figure}{1}  % Set figure counter to 1 for next figure
\addtocounter{figure}{1} % Move counter to 2 so next figure starts at 2

\maketitle
\thispagestyle{empty}
\pagestyle{empty}

\begin{abstract}
Zero-shot open-vocabulary object navigation has progressed rapidly with the emergence of large Vision-Language Models (VLMs) and Large Language Models (LLMs), now widely used as high-level decision-makers instead of end-to-end policies. Although effective, such systems often rely on iterative large-model queries at inference time, introducing latency and computational overhead that limit real-time deployment. 
To address this problem,  we repurpose ray frontiers (R2F), a recently proposed frontier-based exploration paradigm, to develop an LLM-free framework for indoor open-vocabulary object navigation. While ray frontiers were originally used to bias exploration using semantic cues carried along rays, we reinterpret frontier regions as explicit, direction-conditioned semantic hypotheses that serve as navigation goals. Language-aligned features accumulated along out-of-range rays are stored sparsely at frontiers, where each region maintains multiple directional embeddings encoding plausible unseen content. In this way, navigation then reduces to embedding-based frontier scoring and goal tracking within a classical mapping and planning pipeline, eliminating iterative large-model reasoning. We further introduce R2F-VLN, a lightweight extension for free-form language instructions using syntactic parsing and relational verification without additional VLM or LLM components. Experiments in Habitat-sim and on a real robotic platform demonstrate competitive state-of-the-art zero-shot performance with real-time execution, achieving up to 6 times faster runtime than VLM-based alternatives.
\end{abstract}

\section{Introduction}
\label{sec:intro}

Zero-shot open-vocabulary object navigation requires a robot to reach a target specified at test time, either as an object category (\eg ``\emph{find a sink}'')~\cite{majumdar2022zson} or as a free-form language description with attributes and context (\eg ``\emph{go to the round dark wooden table near the staircase}'')~\cite{yu2024vln}, in an unseen indoor environment under a limited time budget. In these settings, success depends on two tightly coupled capabilities: (i) long-horizon exploration to reveal new regions of the environment; and (ii) semantic reasoning to prioritize where to explore based on partial observations.

Frontier-based exploration has long provided a principled geometric solution to the first challenge~\cite{yamauchi1997frontier}. By identifying boundaries between known free space and unexplored regions, frontiers define actionable waypoints that enable systematic coverage of unknown environments. However, classical frontiers are purely geometric constructs: they indicate where exploration can proceed, but not what might lie beyond. In open-vocabulary settings, this limitation becomes critical, since navigation decisions must be conditioned on language-specified targets whose spatial locations are initially unknown. 
Recent work has begun to augment frontier representations with semantic cues~\cite{chen2023not, yokoyama2024vlfm}. In particular, the RayFronts paradigm~\cite{alama2025rayfronts} introduces \emph{ray frontiers}, where language-aligned visual features are propagated along rays extending beyond the sensor range and attached to frontier voxels, providing directional semantic evidence about unexplored space. While originally proposed for outdoor active exploration, this mechanism suggests a promising way to couple frontier-based exploration with open-vocabulary perception.

In parallel, recent approaches to zero-shot semantic navigation rely on large \acp{VLM} and \acp{LLM} to inject semantic priors and repeatedly re-rank candidate exploration targets online~\cite{zhou2025beliefmapnav, sun2025openfrontier, dorbala2023can}. While effective, these systems typically require frequent calls to large models at inference time, introducing latency that complicates real-time robotic deployment~\cite{zhou2023navgpt}. Moreover, semantic reasoning is often derived from global image embeddings, which provide limited directional grounding and therefore weak guidance for frontier-based exploration~\cite{yokoyama2024vlfm, chen2023not}.
These limitations motivate alternative approaches that preserve the efficiency of frontier-based exploration while incorporating language-aligned semantic evidence directly into spatial representations. \looseness-1 

In this work, we revisit frontier-based exploration as the central decision structure for zero-shot navigation. Rather than delegating semantic reasoning to external language models or relying on global semantic embeddings% with limited directional grounding
, we attach open-vocabulary visual evidence directly to frontier regions. To this end, we \emph{repurpose ray frontiers} (\rrf) into an \ac{LLM}-free navigation framework that converts directional semantic cues into explicit navigation goals. Language-aligned features accumulated along out-of-range rays are stored sparsely at frontier regions, where each region maintains multiple direction-conditioned embeddings representing plausible unseen content. Frontier regions are then scored in a shared vision-language embedding space and selected as navigation targets instead of purely geometric waypoints.

Within this framework, we address two tasks: (i) zero-shot open-vocabulary \ac{ObjectNav}, where the agent must locate an arbitrary target object in an unseen environment; and (ii) zero-shot open-vocabulary \ac{VLN}, where the agent follows natural-language instructions involving attributes and spatial relations to reach a goal. By embedding semantic evidence directly into spatial boundaries, \rrf \xspace provides a lightweight, interpretable, and highly reactive framework suitable for robotic deployment. %This design eliminates iterative VLM/LLM querying while preserving the spatial structure of frontier exploration.

To summarize, our main contributions are:
\begin{itemize}
    \item \rrf, a real-time LLM-free and training-free open-vocabulary navigation framework that repurposes ray frontiers into explicit semantic navigation targets.
    \item An embedding-scored frontier selection policy that converts semantic ray frontiers from exploration priors into explicit directional navigation goals while preserving a purely geometric occupancy map.
    \item \rrfv, an extension of the framework to free-form language instructions through lightweight relational verification without additional \ac{VLM} or \ac{LLM} components.
    \item Extensive evaluation in photorealistic simulation and on a real robot, showing real-time execution and up to 6 times faster runtime than VLM-based alternatives.
\end{itemize}

The remainder of this paper is structured as follows. Section~\ref{sec:related_work} reviews related work on frontier-based exploration and object navigation, while Section~\ref{sec:methodology} details the proposed approach. Section~\ref{sec:results} presents the experimental setup and reports results. Finally, Section~\ref{sec:conclusion} concludes the paper and outlines future work.
\section{Related Works}
\label{sec:related_work}
\begin{figure*}[ht!]
    \centering
    \includegraphics[width=\linewidth]{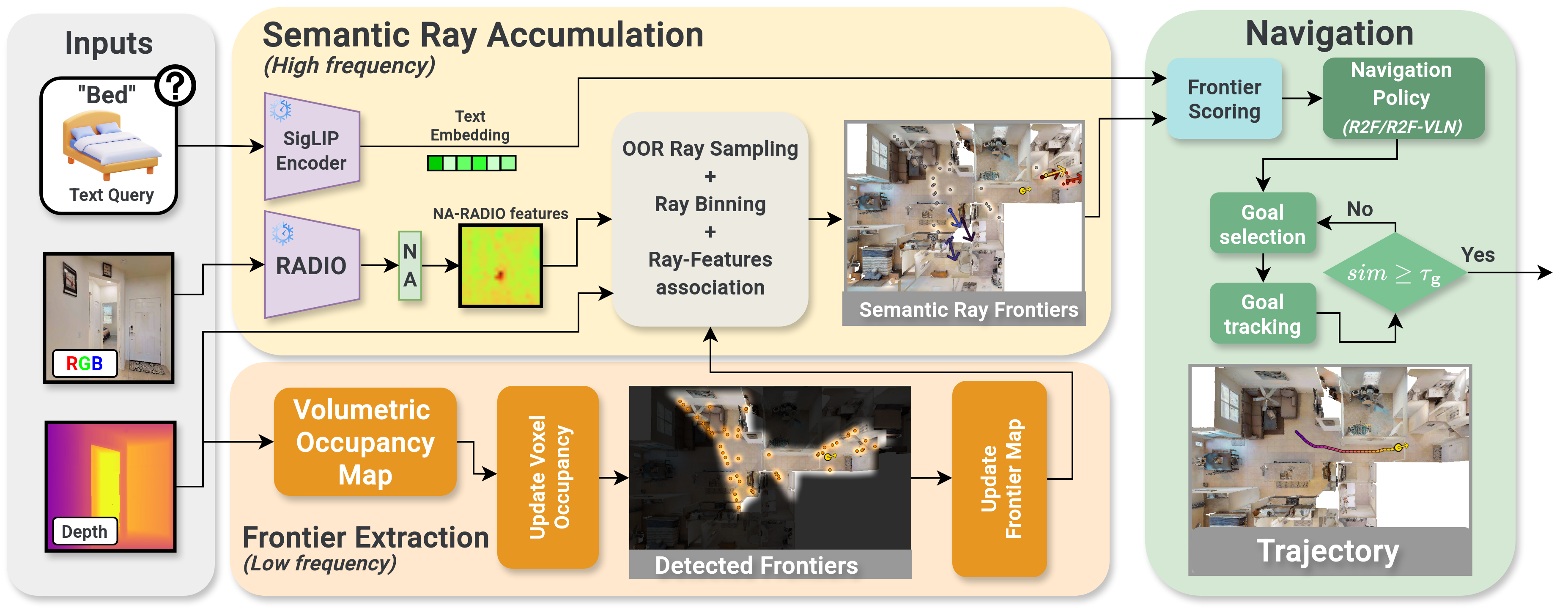}
    \caption{\textbf{\rrf\xspace system overview and execution schedule.} The agent receives RGB-D observations and a text query. RGB images are processed by RADIO with a Neighborhood-Aware attention modification (NA) to produce dense open-vocabulary features (NA-RADIO), while the query is encoded with SigLIP. From out-of-range depth pixels, semantic rays are sampled, binned by direction, and associated with frontier regions, forming Semantic Ray Frontiers that accumulate directional semantic evidence. In parallel, depth updates a volumetric occupancy map, from which frontier regions are periodically recomputed (\emph{low frequency}), while semantic ray accumulation runs continuously (\emph{high frequency}). Frontier regions are scored via cosine similarity with the query embedding, and the navigation policy (\rrf\xspace or \rrfv) selects and tracks the highest-scoring frontier until the goal is detected or exploration continues.}
\label{fig:overview}
\end{figure*}

\boldparagraph{Object-goal navigation} In the \ac{ObjectNav} and \ac{VLN} tasks, an agent must explore previously unseen environments to reach target query objects. Classic approaches build episodic semantic maps to bias exploration towards likely target locations~\cite{chaplot2020object, chang2023goat}. Other methods pair object-centric mapping with utility-driven exploration to search for and reconstruct targets \cite{papatheodorou2023finding}, or propagate semantic priors into geometric exploration to prioritize promising candidates \cite{chen2023not}. Multi-modal embeddings have also been used to enable visual-goal navigation from language-specified targets \cite{majumdar2022zson, gadre2023cows}. More recently, Language Models have been leveraged to provide commonsense priors for long-horizon search, for example by ranking frontiers or subgoals \cite{zhou2023esc,yu2023l3mvn} or combining object probability maps with LLM-derived proximity priors for informative planning \cite{qu2024ippon}.
Yu et al.~\cite{yu2024vln} propose VLN-Game, a zero-shot modular framework that integrates a 3D object-centric map with vision-language features and game-theoretic equilibrium search to improve VLM-based target identification. In contrast, our approach embeds semantics directly into spatial boundaries, enabling a lightweight, prompt-free navigation policy without expensive equilibrium search or repeated \ac{VLM} calls.
%
%
%
%
%
% Frontier-based exploration
%Foundational frontier concept: \cite{yamauchi1997frontier} 
%Frontier selection with added graph/cost cues: \cite{gao2018improved} 
%Fast frontier NBV with entropy/time utility in octree maps: \cite{dai2020fast} 
%Vision-only frontier proposal + info gain prediction: \cite{sun2025frontiernet} 
%Scene-completion-aware information gain: \cite{schmid2022sc} 
%Hierarchical local/global exploration planning: \cite{cao2021tare} 
%Receding-horizon sampling/tree planners: \cite{bircher2018receding} 
%Neural-field uncertainty reduction for NBV: \cite{lee2022uncertainty}

\boldparagraph{Frontier-based exploration} Frontiers mark the boundary between explored and unknown space. Beyond the basic ``closest-frontier'' heuristic ~\cite{yamauchi1997frontier}, many methods incorporate travel cost and topological cues \cite{gao2018improved}, or operate on 3D occupancy maps with sampling-based viewpoint selection guided by utility functions balancing expected information gain and motion cost~\cite{dai2020fast}. Learning-based approaches such as FrontierNet~\cite{sun2025frontiernet} instead predict frontier regions and their expected gain directly from RGB-D observations. Other works exploit higher-level priors, including scene-completion-aware exploration~\cite{schmid2022sc}, hierarchical map representations for multi-scale planning~\cite{cao2021tare}, and receding-horizon exploration using geometric random trees \cite{bircher2018receding}.  

More recently, frontier selection has been augmented with open-vocabulary multi-modal representations that transform geometric frontiers into semantically informed subgoals. VLFM \cite{yokoyama2024vlfm} biases frontier selection using a language-grounded value map derived from a \ac{VLM}, while \acp{LLM} can also rank frontier subgoals as semantic heuristics~\cite{ sun2025openfrontier}. Semantic exploration can also be formulated without explicit frontiers by predicting unseen occupancy and sampling informative viewpoints \cite{tao2022seer}, or by minimizing uncertainty in neural scene representations such as NeRFs \cite{lee2022uncertainty}.  

RayFronts \cite{alama2025rayfronts} is particularly related, as it associates semantic evidence with frontier-like structures. We extend this idea to indoor open-vocabulary navigation by storing direction-conditioned semantic evidence at frontier regions and directly converting it into navigation goals, avoiding iterative \ac{VLM}/\ac{LLM} queries.

\boldparagraph{Semantics-based exploration} 
Several approaches integrate language-aligned semantics directly into spatial representations. VLMaps~\cite{huang2023visual} attach vision-language features to 3D maps to enable language queries over space, while open-vocabulary 3D scene understanding frameworks (\eg OpenScene) provide the perception backbone for retrieving novel concepts~\cite{peng2023openscene}. 
Other approaches treat grounding as an agentic process, verifying spatial and relational constraints over candidates retrieved by open-vocabulary perception \cite{yang2024llm}. 
Open-vocabulary scene graphs \cite{gu2024conceptgraphs, rotondi2025fungraph} and snapshot-based memories \cite{yang20253d} further support exploration and reasoning by maintaining multi-modal context over time. Our approach instead retains standard geometric mapping and planning, while compressing open-vocabulary semantic evidence into a memory coupled to frontier regions.

\section{Methodology}
\label{sec:methodology}

This section is organized as follows. Section \ref{subsec:meth:pf} provides a formal definition of the \ac{ObjectNav} task and \ac{VLN} task. Sections \ref{subsec:meth:semantics}, \ref{subsec:meth:svm}, \ref{subsec:meth:srf} highlight the main modules of our architecture, which can also be observed in Fig.~\ref{fig:overview}. Lastly, Section \ref{subsec:meth:impl} describes the implementation details.

\subsection{Task Formulation}
\label{subsec:meth:pf}
We consider zero-shot open-vocabulary \ac{ObjectNav} and \ac{VLN} in indoor environments as our primary tasks. For brevity, we omit the qualifier zero-shot open-vocabulary in the remainder of this paper.

Given a query $q$ and an initial random position $p_0 \in SE(3)$ inside an environment $e$, these tasks require finding a sequence of actions $\{a_t\}_{t=0}^{T}$ with $a_t \in \mathcal{A}$, that drives the agent to a goal location $g \in \mathcal{G}_q$ consistent with $q$, within a time budget $T_{\max}$. The action space $\mathcal{A}$ includes a dedicated \texttt{STOP} action, which the agent must explicitly execute to declare task completion. At each time step $t$, the agent receives an RGB-D observation $o_t = (I_t, D_t)$, with $I_t \in \mathbb{R}^{H \times W \times 3}$ and $D_t \in \mathbb{R}^{H \times W}$, and updates its pose $p_t \in SE(3)$ by executing an action $a_t$. An episode is considered successful only if the agent issues the \texttt{STOP} action at time $t \leq T_{\max}$ and its planar position is within a distance $\delta$ from the goal set, that is,
\[
\min_{g \in \mathcal{G}_q} \| \bar{p}_t - \bar{g} \|_2 \leq \delta,
\]
where $\bar{p}_t$ denotes the ground-plane projection of $p_t$, and $\bar{g}$ is the projection of $g$ onto the same plane.

In the case of \ac{ObjectNav}, the query specifies an object category $q = c$. The goal set $\mathcal{G}_q = \mathcal{G}_c$ contains the 3D positions of all instances of category $c$ in the environment. The task therefore consists of navigating to any instance of the target category and correctly issuing the \texttt{STOP} action when close to it. In \ac{VLN} instead, the query is a free-form natural language instruction $q \in \mathcal{L}$ that implicitly defines a goal region $\mathcal{G}_q$, which may depend on semantic content, spatial relations, or multi-step directives. Unlike \ac{ObjectNav}, $\mathcal{G}_q$ is not defined at the level of an object category but refers to a specific object instance. Consequently, the target must be inferred from the language instruction before the agent can determine when to terminate. In the zero-shot setting considered in this work, no task-specific training is performed on the evaluation environments and the navigation policy must generalize to unseen scenes and queries.

\subsection{Directional Dense Spatial Semantics}
\label{subsec:meth:semantics}
At the core of our approach lies a joint geometric semantic representation combining a frontier based map with \acp{SRF}. The map encodes free, occupied, and unexplored space, while \acp{SRF} attach direction-aware open-vocabulary evidence to frontier regions. This requires dense features with spatial structure to associate observations with frontier directions.

\begin{figure}[t!]
    \centering
    \includegraphics[width=\linewidth]{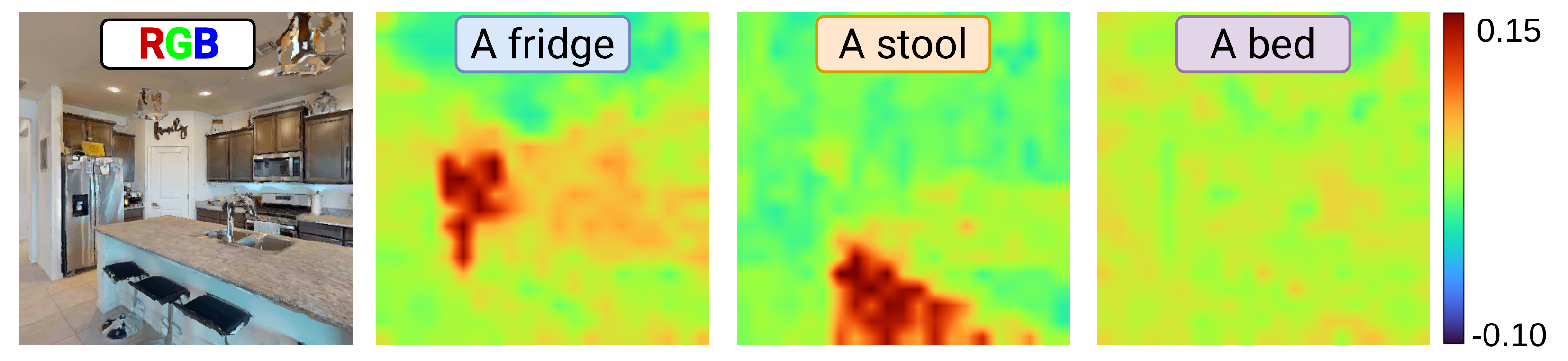}
    \caption{NA-RADIO's feature maps in comparison with different text query. Both the text query embedding and the visual features lie in the SigLIP~\cite{zhai2023sigmoid} embedding space thanks to RADIO's adapter. In practice, we observe that informative similarity values typically lie in the range $[-0.10, 0.15]$.}
    \label{fig:naradio}
\end{figure}

To recover spatially consistent and directionally grounded semantics, we adopt the novel strategy proposed by NACLIP~\cite{hajimiri2025pay} to generate dense patch-level features by modifying the last layer of a \ac{ViT}. 
Being $i$ and $j$ indices over visual tokens (patches) produced by the \ac{ViT} image encoder, standard self-attention computes the pairwise similarity between query and key vectors as part of the multi-head self-attention mechanism. 
In contrast, NACLIP replaces this global similarity formulation with a \emph{neighborhood-aware} attention mechanism that emphasizes local interactions between patch key vectors.
Formally, for each pair of patch tokens $x_i, x_j \in \mathbb{R}^D$, instead of computing the usual query-key dot product
\[
\text{sim}_{ij} = \frac{q_i^\top k_j}{\sqrt{D}},
\]
NACLIP constructs a modified affinity measure based on the \emph{key-key} interaction between patches:
\[
\text{sim}_{ij}^{\text{NACLIP}} = \frac{k_i^\top k_j}{\sqrt{D}} \cdot G\bigl(\|u_i - u_j\|\bigr),
\]
where $G(\cdot)$ is a Gaussian kernel applied to the spatial distance between the patch centers $u_i, u_j \in \mathbb{R}^2$ in the image plane, enforcing locality. By integrating this neighborhood-aware similarity into the self-attention computation, we produce feature maps where each spatial location preserves both semantic alignment to the language query and spatial coherence crucial for dense reasoning. Following~\cite{alama2025rayfronts}, we adopt RADIO~\cite{heinrich2025radiov2} as our backbone, a \ac{ViT} trained via multi-teacher distillation. Specifically, RADIO integrates complementary representations by distilling geometric and texture-aware features from DINO~\cite{caron2021emerging}, vision-language alignment from CLIP~\cite{radford2021learning}, and boundary-aware segmentation cues from SAM~\cite{kirillov2023segment}. To obtain language-aligned dense features, we further leverage RADIO's SigLIP~\cite{zhai2023sigmoid} summary feature adapter, which projects spatial patch-level features into the SigLIP \texttt{[CLS]}-token embedding space. This projection preserves spatial structure while aligning visual features with the text embedding space, enabling direct cosine similarity with the query and yielding a spatially consistent, language-aligned dense feature map for open-vocabulary reasoning. Qualitative results of Neighborhood-Aware RADIO (NA-RADIO) feature maps can be observed in Fig.~\ref{fig:naradio}.

\subsection{Volumetric Frontier Representation}
\label{subsec:meth:svm}

We maintain a probabilistic volumetric representation of the environment that models free, occupied, and unknown space, following standard wavelet-based occupancy mapping approaches~\cite{reijgwart2023efficient}. Occupancy is represented in log-odds form,
$$
\ell(x) = \log \frac{P(m(x)=1)}{1 - P(m(x)=1)},
$$
where $m(x)=1$ denotes occupancy at location $x \in \mathbb{R}^3$. Given a posed RGB-D observation $o_t$, each pixel induces a 3D ray in the world frame. Under a standard inverse sensor model, free-space evidence is accumulated along the ray up to the measured depth, while occupied evidence is integrated at the surface location. This recursive Bayesian update enables efficient incremental mapping while preserving uncertainty in unobserved regions.

To extract navigational structure, the map is periodically queried and spatial locations are classified according to their occupancy belief. Given the log-odds field $\ell(x)$, we define free space as $\mathcal{F} = \{ x \mid \ell(x) < \tau_{\text{free}} \}$ and occupied space as $\mathcal{O} = \{ x \mid \ell(x) > \tau_{\text{occ}} \}$, where $\tau_{\text{free}}$ and $\tau_{\text{occ}}$ are occupancy thresholds; locations satisfying $\tau_{\text{free}} \le \ell(x) \le \tau_{\text{occ}}$ remain unknown. A spatial location $x \in \mathcal{F}$ is defined as a \emph{frontier} if: (i) it satisfies $y_{\min} \le y(x) \le y_{\max}$, restricting frontiers to the navigable height band of the agent; (ii) the number of 6-connected\footnote{6-connectivity denotes face-adjacency in the 3D voxel grid.} neighbors belonging to unknown space is at least $k_u$, and (iii) the number of 6-connected neighbors belonging to $\mathcal{F}$ is at least $k_f$. These voxels lie at the boundary between explored and unexplored regions and therefore define candidate exploration targets, following the classical frontier-based exploration paradigm~\cite{yamauchi1997frontier}.

Frontier voxels are subsequently aggregated into spatially coherent regions, each represented by its centroid and associated boundary elements. 
%This abstraction provides a stable geometric scaffold that summarizes where exploration can progress, while decoupling local voxel fluctuations from higher-level planning decision.

\subsection{Semantic Ray Frontiers}
\label{subsec:meth:srf}

While the frontier-based geometric map identifies where exploration can progress, it does not encode what may lie beyond currently observed space. To address this limitation, we build upon the RayFronts~\cite{alama2025rayfronts} paradigm and adapt it to \ac{ObjectNav} and \ac{VLN}. RayFronts reinterpret classical frontiers as directional carriers of semantic evidence gathered along rays extending into unexplored regions. Frontier regions therefore act as sparse, direction-conditioned semantic hypotheses about unseen space, while geometric occupancy remains handled by the volumetric map described in Section~\ref{subsec:meth:svm}\looseness-1.

\boldparagraph{Out-of-range semantic rays}
Depth sensing is truncated at a maximum range $r_{\max}$. Pixels satisfying $D_t(u,v) \ge r_{\max}$ correspond to camera rays that are not terminated by an observed surface and therefore propagate into unobserved space. These \ac{OOR} pixels define candidate semantic rays directed toward unexplored regions. To mitigate feature contamination near depth discontinuities, the \ac{OOR} mask is eroded in the image plane prior to further processing. From the remaining pixels, a bounded subset is selected to ensure a controlled number of rays per frame. For each selected pixel, the corresponding world-frame unit direction $\mathbf{d} \in \mathbb{R}^3$ is computed from the current camera pose $p_t$. The associated dense visual feature is retrieved from the NA-RADIO patch-level feature map $\mathbf{f}$, which provides spatially resolved embeddings in a shared vision-language representation space. These normalized embeddings serve as the semantic descriptors carried by the \ac{OOR} rays. 

\boldparagraph{Ray-to-frontier association}
Frontier voxels are clustered into spatially coherent regions with centroids $c_m \in \mathbb{R}^3$. For a camera position $p_t$, define the relative vector $\mathbf{v}_m = c_m - p_t$. Given a ray with unit direction $\mathbf{d}$, its parametric form is $\mathbf{r}(\lambda) = p_t + \lambda \mathbf{d}$, with $\lambda \ge 0$.
Following the geometric consistency principles of RayFronts, region $m$ is considered compatible with a ray of unit direction $\mathbf{d}$ if the following conditions hold:
(i) the region lies in front of the ray, $\mathbf{d}^\top \mathbf{v}_m > 0$;
(ii) the perpendicular distance between the ray and the centroid,
$
d_{\perp}(m)
=
\left\| \mathbf{v}_m - (\mathbf{d}^\top \mathbf{v}_m)\mathbf{d} \right\|,
$
is below a fixed tolerance $\tau_{\perp}$;
and (iii) the radial distance
$
\|\mathbf{v}_m\|
$
remains below a maximum association range $\tau_r$.
Among all regions satisfying these constraints, the ray is assigned to the region minimizing a normalized geometric cost that combines perpendicular deviation and radial distance. The corresponding feature vector $\mathbf{f}$ carried by the ray is then accumulated into the selected region using a weighted running average. 

\boldparagraph{Storing semantics efficiently} In contrast to RayFronts, semantic evidence is not fused volumetrically into the occupancy grid and no ray re-casting is performed. Instead, semantic information is stored exclusively at frontier regions. Each region maintains a set of discrete direction bins obtained by partitioning spherical angles with a fixed angular resolution, and for each bin $b$ stores a weighted average feature vector $\mathbf{f}_{m,b}$. This separation preserves a purely geometric occupancy map while allowing frontier regions to encode multiple semantic interpretations of the same spatial boundary that are direction-dependent.

\boldparagraph{Execution schedule}
Semantic ray accumulation is performed at every timestep, integrating dense visual features into frontier regions online. Frontier extraction and region synchronization, instead, are executed periodically every $N_{map}$ steps by querying the volumetric map. This two-rate execution scheme balances high-frequency semantic updates with more expensive geometric frontier recomputation. More details in Section~\ref{subsec:meth:impl}.

\subsection{Navigation Policy}
\label{subsec:meth:policy}

In contrast to RayFronts, where frontier semantics mainly bias exploration, we treat semantic ray frontiers as \emph{explicit navigation goals}. Each frontier region, enriched with direction-conditioned embeddings, is directly used as a waypoint for global path planning. 

\boldparagraph{Semantic scoring of frontier regions}
Given the normalized query embedding $\mathbf{t}_q \in \mathbb{R}^D$, semantic alignment is evaluated via cosine similarity with the region features. For each region $m$ and direction bin $b$, we compute $s_{m,b} = \mathbf{f}_{m,b}^\top \mathbf{t}_q$, since both embeddings are $\ell_2$-normalized. The semantic score of region $m$ is defined as $S_m = \max_b \; s_{m,b}$, selecting the most language-consistent directional hypothesis associated with that frontier. The maximizing bin determines the preferred exploration direction. Regions are ranked in descending order of $S_m$, yielding a prioritized set of semantic navigation candidates.

\boldparagraph{\rrf \xspace policy} The overall policy operates as a two-mode state machine alternating between \emph{goal selection} and \emph{goal tracking}. 
During goal selection, the agent updates the geometric map, extracts the current frontier regions, and recomputes semantic scores $\{S_m\}$. % using the current query embedding. 
If at least one region has accumulated semantic evidence, the region with maximal score is selected as the navigation target. Its centroid is projected onto the navigable surface prior to path planning to ensure feasibility. %In the absence of semantically scored regions, the agent falls back to geometric exploration by selecting the nearest unexplored frontier. If no valid frontier exists, the episode terminates.
If no region contains semantic evidence, the nearest unexplored frontier is selected; if none exists, the episode terminates.

During goal tracking, the agent follows the planned waypoint sequence using a reactive local controller. Upon reaching the selected frontier, the corresponding region is invalidated to prevent reselection, and the system returns to goal selection while incorporating newly observed semantic evidence. 
If navigation becomes infeasible or persistent stalling is detected, a new target is selected.

In parallel, a semantic goal detector runs at every timestep to capture direct visual evidence of the queried object. From the spatial feature map of the current observation, the highest query-consistent response across all spatial locations is extracted. 
A detection threshold $\tau_g$ determines goal presence, and the object is declared found once this response exceeds $\tau_g$ for a predefined number of consecutive frames $N_{cons}$. 
The spatial location corresponding to the highest-scoring feature is then reprojected into world coordinates to produce a 3D goal hypothesis, and the agent performs a final local approach toward this point to complete the task before firing \texttt{STOP}.\looseness-1

\boldparagraph{\rrfv \xspace policy} While the base \rrf \xspace policy performs category-level grounding through embedding alignment, free-form language instructions additionally require enforcing relational and compositional consistency. We therefore extend the same geometric-semantic framework with a lightweight \ac{NLP} stage that introduces relational verification on top of category grounding, without altering the underlying frontier-based navigation mechanism.\looseness-1

Given a language query $q$, the instruction is parsed into a target phrase and a set of landmark objects. Noun chunks and their modifiers are extracted from the dependency tree, filtering out pronouns and directional terms. The target is selected as the syntactic subject when available, otherwise as the first noun phrase preceding a spatial relation. The remaining relational noun phrases are treated as landmarks.

The target phrase, formed by concatenating attributes and head noun, is encoded to produce the query embedding $\mathbf{t}_q$ used for frontier scoring and dense detection. For each landmark, WordNet-based~\cite{miller1995wordnet} lexical variants are generated and filtered in the embedding space using a similarity threshold $\tau_{{syn}}$, retaining at most $K_{\text{syn}}$ embeddings.

When a candidate detection is triggered after $N_{{confirm}}$ consecutive frames above $\tau_g$, the agent performs a short rotational sweep and accumulates the maximum similarity between observed features and each landmark embedding. 
A detection is confirmed if at least one landmark exceeds a threshold $\tau_{\ell}$, otherwise, it is rejected and exploration continues.
This procedure enables relational grounding without additional \acp{VLM}, still preserving real-time performance.

\subsection{Implementation Details}
\label{subsec:meth:impl}

All experiments are conducted in Habitat-sim~\cite{habitat19iccv} with a discrete action space consisting of \texttt{move\_forward} (0.25\,m), \texttt{turn\_left}/\texttt{turn\_right} (15$^\circ$ yaw), and \texttt{STOP}. The agent receives RGB-D observations, with depth clipped to $r_{\max}=3.5$\,m and a camera height of 1.25\,m. 
The maximum timestep budget is $T_{\max}=1000$ steps per episode, and task success is defined with tolerance $\delta=1.5$\,m as introduced in Section~\ref{subsec:meth:pf}. Path planning assumes access to the environment \emph{navmesh} provided by the simulator.
Geometric mapping is implemented with WaveMap~\cite{reijgwart2023efficient} using a hashed chunked wavelet octree, with voxel size 0.1\,m and minimum cell width 0.05\,m. 

Following the execution schedule described in Section~\ref{subsec:meth:srf}, semantic ray accumulation and dense feature extraction are performed at every timestep. Instead, frontier extraction, occupancy querying, and region synchronization are executed every $N_{\text{map}}=5$ steps. This periodic update amortizes the cost of volumetric querying while maintaining high-frequency semantic integration.

In all experiments, we use a merge radius of 0.8\,m, association limits $\tau_r=14.0$\,m and $\tau_{\perp}=1.0$\,m, an angular bin size of 30$^\circ$, an invalidation radius of 1.0\,m, a visited filter distance of 2.0\,m, and frontier extraction thresholds $k_u=3$ unknown neighbors and $k_f=1$ free neighbor. Dense goal detection requires $N_{cons}=3$ consecutive frames above the threshold $\tau_g=0.14$ to declare an object found. In \rrfv\xspace, a candidate is triggered after $N_{confirm}=3$ consecutive frames and subsequently verified through landmark confirmation with threshold $\tau_\ell=0.11$. Synonym filtering relies on $\tau_{\text{syn}}=0.60$ with $K_{syn}=5$.
For the complete list of hyperparameters and configuration details, we refer to the project repository.
\label{subsec:meth:impl}

\section{Experimental Results}
\label{sec:results}

\subsection{Dataset}
We evaluate our approach on the \ac{ObjectNav} and \ac{VLN} navigation tasks defined in the VLN-Game benchmark~\cite{yu2024vln}. Both tasks are built upon photorealistic 3D environments from the Habitat-Matterport 3D (HM3D) dataset~\cite{ramakrishnan2021habitat}, which provides large-scale indoor scenes with accurate geometry and semantic annotations. GOAT-Bench \cite{khanna2024goat}, a benchmark for multi-modal lifelong navigation that adds language descriptions as goals and evaluates general-purpose navigation systems across diverse settings, is used to evaluate the pipeline and the comparison pipelines.

For \ac{ObjectNav}, we use the \texttt{objectnav\_hm3d\_v1} split, and for \ac{VLN} the \texttt{vlobjectnav\_hm3d\_v4} split adapted by VLN-Game from GOAT-Bench~\cite{khanna2024goat}.

We extract fixed evaluation subsets to ensure controlled and reproducible comparisons. The \ac{ObjectNav} subset contains 60 episodes across 10 scenes (6 per scene) from the \texttt{objectnav\_hm3d\_v1} validation split, while the \ac{VLN} subset contains 50 episodes across 10 scenes (5 per scene) from \texttt{vlobjectnav\_hm3d\_v4}. The subsets are available in the project repository.

Each \ac{VLN} episode specifies a start position, a target object instance identified by name, its 3D goal location, and a natural language instruction describing the target relative to surrounding landmarks, for example \emph{``king size bed located near the chest drawer, painting, curtain, and pillow''}. In contrast, each \ac{ObjectNav} episode provides a start position, a target object category, and the 3D locations of all goal instances belonging to that category within the scene.

\subsection{Baselines}
We compare our method against baselines representing common approaches to language-guided navigation: VLM-based semantic reasoning (VLN-Game), memory-based scene modeling (3D-Mem), and frontier exploration augmented with language models (VLFM, OpenFrontier).

\textbf{VLN-Game}~\cite{yu2024vln} is a state-of-the-art framework for zero-shot \ac{VLN} that performs %semantic grounding through 
iterative reasoning with a \ac{VLM}. It constructs a semantic map of candidate objects %from visual observations
and formulates navigation as a game-theoretic selection process aligning detected objects with the language instruction.

\textbf{3D-Mem}~\cite{yang20253d} represents explored regions using \emph{Memory Snapshots}, multi-view images capturing clusters of co-visible objects and their context. Exploration is guided by \emph{Frontier Snapshots}, which store visual glimpses of unexplored areas, while a \ac{VLM} reasons over both memory and frontier snapshots to select navigation actions.

\textbf{VLFM}~\cite{yokoyama2024vlfm} combines geometric frontier exploration with vision-language grounding. The system detects frontiers from an occupancy map and uses a pre-trained \ac{VLM} to score nearby observations with respect to the target query, prioritizing frontiers likely to lead to the goal.

\textbf{OpenFrontier}~\cite{sun2025openfrontier} extends FrontierNet~\cite{sun2025frontiernet}, which selects frontiers with high %predicted geometric 
information gain from RGB observations. OpenFrontier re-scores these frontiers using a \ac{VLM} to prioritize regions likely to contain the target object.

For VLN-Game, 3D-Mem, and VLFM we use the authors’ public implementations. As no official code is available for OpenFrontier, we extend the public FrontierNet implementation to reproduce the method.

\subsection{Metrics}

\begin{comment}
The experimental evaluation focuses on three metrics: Success Rate (SR), Success weighted by Path Length (SPL), and Execution time.

\begin{itemize}
    \item \textbf{Success Rate (SR):} Measures weather the agent successfully reaches the target object. An episode is considered successful if the final geodesic distance to the goal computed via the navmesh rather than the Euclidean planar distance lies within $\delta$ (1.5 meters in our case)
    \item \textbf{Success weighted by Path Length (SPL):} Evaluates the efficiency of the agent's trajectory by comparing the actual path taken against the shortest possible geodesic path from the starting position to the goal
    \item \textbf{Execution Time:} Measures the time required to complete the episode, reflecting the computational and temporal efficiency of the navigation pipeline
\end{itemize}
\end{comment}

The experimental evaluation considers three metrics: \emph{\ac{SR}}, \emph{\ac{SPL}}, and \emph{Execution Time}. 
For each episode $i \in \{1, \dots, N\}$, define the success indicator $S_i \in \{0,1\}$, where $S_i = 1$ if the final distance to the goal is below the threshold $\delta = 1.5$ meters, and $S_i = 0$ otherwise. The \ac{SR} is defined as
$$
\text{SR} = \frac{1}{N} \sum_{i=1}^{N} S_i.
$$

Let $L_i$ denote the executed path length and $L_i^*$ the optimal shortest-path length for episode $i$. The \ac{SPL} is defined as
$$
\text{SPL} = \frac{1}{N} \sum_{i=1}^{N} S_i \frac{L_i^*}{max(L_i,L_i^*)},
$$
so that only successful episodes contribute and longer trajectories are penalized proportionally.

Execution Time corresponds to the total time required to complete an episode, reflecting the overall computational efficiency of the navigation pipeline. To ensure fair comparison, all baselines have been run on the same machine under equal conditions.

\begin{table}[t]
\centering
\caption{Comparison across \ac{ObjectNav} and \ac{VLN} tasks with the three baselines. We report \emph{\ac{SR}}, \emph{\ac{SPL}}, and Execution Time ($t$). Bold is best and underlined is second-best.}
\label{tab:results_table}
\resizebox{0.49\textwidth}{!}{
\begin{tabular}{c c c c c}
\toprule
Task & Method & \ac{SR} (\%) & \ac{SPL} (\%) & $t$ (s) \\
\midrule

\multirow{5}{1em}{\rotatebox{90}{ObjectNav}}
& VLN-Game~\cite{yu2024vln}                  & \underline{76.7}   & \underline{28.0}        & 122.0\\
& 3D-MEM~\cite{yang20253d}                   & 26.7   & 15.9      & \underline{61.2}    \\ 
& VLFM~\cite{yokoyama2024vlfm}               & 40.0   & 7.71       & 84.2    \\ % final fixed results
& OpenFrontier~\cite{sun2025openfrontier}    & 23.3   & 7.12       & 245.0    \\ % final fixed results
& \rrf \xspace(ours)                & \textbf{78.3}   & \textbf{29.6}     & \textbf{32.7}        \\

\midrule
\multirow{5}{1em}{\rotatebox{90}{VLN}}
& VLN-Game~\cite{yu2024vln}                 & \textbf{43.7}     & \textbf{22.7}     & 504.0      \\
& 3D-MEM~\cite{yang20253d}                  & 18.0      & 12.3     & \underline{57.3}  \\
& VLFM~\cite{yokoyama2024vlfm}              & \underline{28.0}      & 6.63      & 133.0   \\ % final fixed results
& OpenFrontier~\cite{sun2025openfrontier}   & 9.7      & 2.12      & 363.0   \\ % final fixed results
& \rrfv \xspace(ours)         & \underline{28.0}     & \underline{13.94}     & \textbf{40.3}        \\

\bottomrule
\end{tabular}
}
\end{table}

\subsection{Experiments}
\label{subsec:results:exp}
Results of our experiments are reported in Table~\ref{tab:results_table}. In \ac{ObjectNav}, \rrf \xspace consistently outperforms all other methods across all three metrics, highlighting the effectiveness of our approach. Overall, our method is twice as fast as the second-fastest baseline while achieving 3 times higher accuracy, and it is 4 times faster than the second most accurate method, VLN-Game. In contrast, \rrfv \xspace achieves the second-best performance on \ac{VLN}. While it is about 6 times faster than VLN-Game due to avoiding repeated \ac{VLM} calls, its performance is limited by a reduced compositional understanding of the scene. Failure analysis indicates that the main errors arise from false positives, where objects with landmarks similar to the query are detected but appear in a different configuration. Such cases are typically resolved by a \ac{VLM}, whereas they remain challenging for \rrfv.
We use the synthetic dataset for controlled benchmarking, but the final validation is carried out on a real robotic platform, where the proposed approach demonstrates robust performance.

\subsection{Real-world Validation}

\label{subsec:results:real-world}
\begin{figure}
    \centering
    \includegraphics[width=1.0\linewidth]{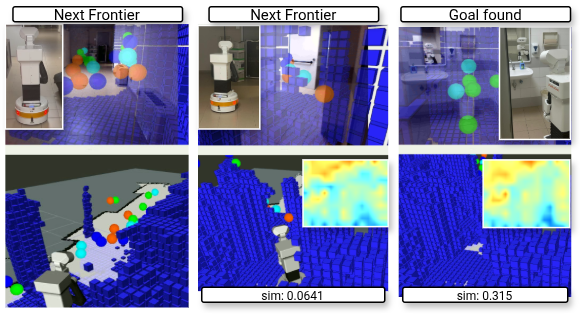}
    \caption{Pipeline execution on a real robot for the ``Find a sink'' goal. The robot navigates between the boundaries based on their semantic value until it reaches the goal.}
    \label{fig:real_robot}
\end{figure}

We validated our approach on a real-world scenario, by implementing it as a ROS package and deploying it on a TIAGo robot. The robot was asked to find \emph{``a sink''} starting from the corridor, and it navigated through the basement and laboratories to reach the designated object in the bathroom, as seen in \ref{fig:real_robot}.
Our software runs on a laptop equipped with an Intel Core 9 Ultra 185H, 32 GB of RAM, and an NVIDIA GeForce 4070 with 8 GB of VRAM. It achieves an average inference rate of 25 Hz, enabling real-time image processing.
\section{Conclusion}
\label{sec:conclusion}

In this paper we introduced \rrf, a real-time, \ac{LLM}-free framework for zero-shot open-vocabulary navigation that repurposes ray frontiers as semantic navigation hypotheses. By attaching language-aligned visual evidence directly to frontier regions, the method converts exploration boundaries into explicit navigation targets while preserving a standard geometric mapping and planning pipeline.
The resulting policy combines open-vocabulary perception with frontier-based exploration. % in a lightweight and reactive manner. 
We further extend the approach to free-form language instructions through \rrfv, which adds a simple relational verification stage without introducing additional large models. Experiments show that \rrf achieves competitive performance on \ac{ObjectNav} while significantly reducing execution time compared to methods relying on iterative \acp{VLM} or \acp{LLM}. In the more challenging \ac{VLN} setting, the approach remains faster than alternatives based on large models, though performance is limited by weaker compositional reasoning. Real-world experiments on a mobile robot further demonstrate reliable real-time operation. %Overall, embedding open-vocabulary semantics directly into frontier representations provides an efficient alternative to large-model reasoning for embodied navigation.
Future work will focus on improving compositional grounding and extending the framework to more complex environments and long-horizon tasks.
%commented for submission
\section*{Acknowledgements}
This work has been carried out while Emanuele Musumeci, Michele Brienza and Francesco Argenziano were enrolled in the Italian National Doctorate on Artificial Intelligence run by Sapienza University of Rome. Michele Brienza is funded by the European Union - Next Generation EU, Mission I.4.1 Borse PNRR Pubblica Amministrazione (Missione 4) Component 1 CUP B53C23003540006. This work has been partially supported by PNRR MUR project PE0000013-FAIR.

\bibliographystyle{IEEEtran}
\bibliography{references.bib}

@article{chaplot2020object,
  title={Object goal navigation using goal-oriented semantic exploration},
  author={Chaplot, Devendra Singh and Gandhi, Dhiraj Prakashchand and Gupta, Abhinav and Salakhutdinov, Russ R},
  journal={Advances in Neural Information Processing Systems},
  volume={33},
  pages={4247--4258},
  year={2020}
}

@inproceedings{yamauchi1997frontier,
  title={A frontier-based approach for autonomous exploration},
  author={Yamauchi, Brian},
  booktitle={Proceedings 1997 IEEE International Symposium on Computational Intelligence in Robotics and Automation CIRA'97.'Towards New Computational Principles for Robotics and Automation'},
  pages={146--151},
  year={1997},
  organization={IEEE}
}

@article{sun2025frontiernet,
  title={Frontiernet: Learning visual cues to explore},
  author={Sun, Boyang and Chen, Hanzhi and Leutenegger, Stefan and Cadena, Cesar and Pollefeys, Marc and Blum, Hermann},
  journal={IEEE Robotics and Automation Letters},
  year={2025},
  publisher={IEEE}
}

@inproceedings{dai2020fast,
  title={Fast frontier-based information-driven autonomous exploration with an mav},
  author={Dai, Anna and Papatheodorou, Sotiris and Funk, Nils and Tzoumanikas, Dimos and Leutenegger, Stefan},
  booktitle={2020 IEEE international conference on robotics and automation (ICRA)},
  pages={9570--9576},
  year={2020},
  organization={IEEE}
}

@inproceedings{gao2018improved,
  title={An improved frontier-based approach for autonomous exploration},
  author={Gao, Wenchao and Booker, Matthew and Adiwahono, Albertus and Yuan, Miaolong and Wang, Jiadong and Yun, Yau Wei},
  booktitle={2018 15th international conference on control, automation, robotics and vision (ICARCV)},
  pages={292--297},
  year={2018},
  organization={IEEE}
}

@article{schmid2022sc,
  title={Sc-explorer: Incremental 3d scene completion for safe and efficient exploration mapping and planning},
  author={Schmid, Lukas and Cheema, Mansoor Nasir and Reijgwart, Victor and Siegwart, Roland and Tombari, Federico and Cadena, Cesar},
  journal={arXiv preprint arXiv:2208.08307},
  year={2022}
}

@article{papatheodorou2023finding,
  title={Finding things in the unknown: Semantic object-centric exploration with an mav},
  author={Papatheodorou, Sotiris and Funk, Nils and Tzoumanikas, Dimos and Choi, Christopher and Xu, Binbin and Leutenegger, Stefan},
  journal={arXiv preprint arXiv:2302.14569},
  year={2023}
}

@article{tao2022seer,
  title={Seer: Safe efficient exploration for aerial robots using learning to predict information gain},
  author={Tao, Yuezhan and Wu, Yuwei and Li, Beiming and Cladera, Fernando and Zhou, Alex and Thakur, Dinesh and Kumar, Vijay},
  journal={arXiv preprint arXiv:2209.11034},
  year={2022}
}

@inproceedings{yokoyama2024vlfm,
  title={Vlfm: Vision-language frontier maps for zero-shot semantic navigation},
  author={Yokoyama, Naoki and Ha, Sehoon and Batra, Dhruv and Wang, Jiuguang and Bucher, Bernadette},
  booktitle={2024 IEEE International Conference on Robotics and Automation (ICRA)},
  pages={42--48},
  year={2024},
  organization={IEEE}
}

@article{chen2023not,
  title={How to not train your dragon: Training-free embodied object goal navigation with semantic frontiers},
  author={Chen, Junting and Li, Guohao and Kumar, Suryansh and Ghanem, Bernard and Yu, Fisher},
  journal={arXiv preprint arXiv:2305.16925},
  year={2023}
}

@article{qu2024ippon,
  title={Ippon: Common sense guided informative path planning for object goal navigation},
  author={Qu, Kaixian and Tan, Jie and Zhang, Tingnan and Xia, Fei and Cadena, Cesar and Hutter, Marco},
  journal={arXiv preprint arXiv:2410.19697},
  year={2024}
}

@inproceedings{cao2021tare,
  title={TARE: A hierarchical framework for efficiently exploring complex 3D environments.},
  author={Cao, Chao and Zhu, Hongbiao and Choset, Howie and Zhang, Ji},
  booktitle={Robotics: Science and Systems},
  volume={5},
  pages={2},
  year={2021}
}

@article{bircher2018receding,
  title={Receding horizon path planning for 3D exploration and surface inspection},
  author={Bircher, Andreas and Kamel, Mina and Alexis, Kostas and Oleynikova, Helen and Siegwart, Roland},
  journal={Autonomous Robots},
  volume={42},
  number={2},
  pages={291--306},
  year={2018},
  publisher={Springer}
}

@article{lee2022uncertainty,
  title={Uncertainty guided policy for active robotic 3d reconstruction using neural radiance fields},
  author={Lee, Soomin and Chen, Le and Wang, Jiahao and Liniger, Alexander and Kumar, Suryansh and Yu, Fisher},
  journal={IEEE Robotics and Automation Letters},
  volume={7},
  number={4},
  pages={12070--12077},
  year={2022},
  publisher={IEEE}
}

@article{chang2023goat,
  title={Goat: Go to any thing},
  author={Chang, Matthew and Gervet, Theophile and Khanna, Mukul and Yenamandra, Sriram and Shah, Dhruv and Min, So Yeon and Shah, Kavit and Paxton, Chris and Gupta, Saurabh and Batra, Dhruv and others},
  journal={arXiv preprint arXiv:2311.06430},
  year={2023}
}

@inproceedings{khanna2024goat,
  title={Goat-bench: A benchmark for multi-modal lifelong navigation},
  author={Khanna, Mukul and Ramrakhya, Ram and Chhablani, Gunjan and Yenamandra, Sriram and Gervet, Theophile and Chang, Matthew and Kira, Zsolt and Chaplot, Devendra Singh and Batra, Dhruv and Mottaghi, Roozbeh},
  booktitle={Proceedings of the IEEE/CVF Conference on Computer Vision and Pattern Recognition},
  pages={16373--16383},
  year={2024}
}

@article{yu2024vln,
  title={Vln-game: Vision-language equilibrium search for zero-shot semantic navigation},
  author={Yu, Bangguo and Liu, Yuzhen and Han, Lei and Kasaei, Hamidreza and Li, Tingguang and Cao, Ming},
  journal={arXiv preprint arXiv:2411.11609},
  year={2024}
}

@article{ramakrishnan2021habitat,
  title={Habitat-matterport 3d dataset (hm3d): 1000 large-scale 3d environments for embodied ai},
  author={Ramakrishnan, Santhosh K and Gokaslan, Aaron and Wijmans, Erik and Maksymets, Oleksandr and Clegg, Alex and Turner, John and Undersander, Eric and Galuba, Wojciech and Westbury, Andrew and Chang, Angel X and others},
  journal={arXiv preprint arXiv:2109.08238},
  year={2021}
}

@inproceedings{yu2023l3mvn,
  title={L3mvn: Leveraging large language models for visual target navigation},
  author={Yu, Bangguo and Kasaei, Hamidreza and Cao, Ming},
  booktitle={2023 IEEE/RSJ International Conference on Intelligent Robots and Systems (IROS)},
  pages={3554--3560},
  year={2023},
  organization={IEEE}
}

@inproceedings{gadre2023cows,
  title={Cows on pasture: Baselines and benchmarks for language-driven zero-shot object navigation},
  author={Gadre, Samir Yitzhak and Wortsman, Mitchell and Ilharco, Gabriel and Schmidt, Ludwig and Song, Shuran},
  booktitle={Proceedings of the IEEE/CVF Conference on Computer Vision and Pattern Recognition},
  pages={23171--23181},
  year={2023}
}

@article{majumdar2022zson,
  title={Zson: Zero-shot object-goal navigation using multimodal goal embeddings},
  author={Majumdar, Arjun and Aggarwal, Gunjan and Devnani, Bhavika and Hoffman, Judy and Batra, Dhruv},
  journal={Advances in Neural Information Processing Systems},
  volume={35},
  pages={32340--32352},
  year={2022}
}

@inproceedings{zhou2023esc,
  title={Esc: Exploration with soft commonsense constraints for zero-shot object navigation},
  author={Zhou, Kaiwen and Zheng, Kaizhi and Pryor, Connor and Shen, Yilin and Jin, Hongxia and Getoor, Lise and Wang, Xin Eric},
  booktitle={International Conference on Machine Learning},
  pages={42829--42842},
  year={2023},
  organization={PMLR}
}

@inproceedings{huang2023visual,
  title={Visual language maps for robot navigation},
  author={Huang, Chenguang and Mees, Oier and Zeng, Andy and Burgard, Wolfram},
  booktitle={2023 IEEE International Conference on Robotics and Automation (ICRA)},
  pages={10608--10615},
  year={2023},
  organization={IEEE}
}

@inproceedings{peng2023openscene,
  title={Openscene: 3d scene understanding with open vocabularies},
  author={Peng, Songyou and Genova, Kyle and Jiang, Chiyu and Tagliasacchi, Andrea and Pollefeys, Marc and Funkhouser, Thomas and others},
  booktitle={Proceedings of the IEEE/CVF conference on computer vision and pattern recognition},
  pages={815--824},
  year={2023}
}

@inproceedings{gu2024conceptgraphs,
  title={Conceptgraphs: Open-vocabulary 3d scene graphs for perception and planning},
  author={Gu, Qiao and Kuwajerwala, Ali and Morin, Sacha and Jatavallabhula, Krishna Murthy and Sen, Bipasha and Agarwal, Aditya and Rivera, Corban and Paul, William and Ellis, Kirsty and Chellappa, Rama and others},
  booktitle={2024 IEEE International Conference on Robotics and Automation (ICRA)},
  pages={5021--5028},
  year={2024},
  organization={IEEE}
}

@inproceedings{yang2024llm,
  title={Llm-grounder: Open-vocabulary 3d visual grounding with large language model as an agent},
  author={Yang, Jianing and Chen, Xuweiyi and Qian, Shengyi and Madaan, Nikhil and Iyengar, Madhavan and Fouhey, David F and Chai, Joyce},
  booktitle={2024 IEEE International Conference on Robotics and Automation (ICRA)},
  pages={7694--7701},
  year={2024},
  organization={IEEE}
}

@article{zhou2025beliefmapnav,
  title={Beliefmapnav: 3d voxel-based belief map for zero-shot object navigation},
  author={Zhou, Zibo and Hu, Yue and Zhang, Lingkai and Li, Zonglin and Chen, Siheng},
  journal={arXiv preprint arXiv:2506.06487},
  year={2025}
}

@inproceedings{alama2025rayfronts,
  title={RayFronts: Open-set semantic ray frontiers for online scene understanding and exploration},
  author={Alama, Omar and Bhattacharya, Avigyan and He, Haoyang and Kim, Seungchan and Qiu, Yuheng and Wang, Wenshan and Ho, Cherie and Keetha, Nikhil and Scherer, Sebastian},
  booktitle={2025 IEEE/RSJ International Conference on Intelligent Robots and Systems (IROS)},
  pages={5930--5937},
  year={2025},
  organization={IEEE}
}

@inproceedings{hajimiri2025pay,
  title={Pay attention to your neighbours: Training-free open-vocabulary semantic segmentation},
  author={Hajimiri, Sina and Ben Ayed, Ismail and Dolz, Jose},
  booktitle={Proceedings of the Winter Conference on Applications of Computer Vision},
  pages={5061--5071},
  year={2025}
}

@inproceedings{heinrich2025radiov2,
  title={Radiov2. 5: Improved baselines for agglomerative vision foundation models},
  author={Heinrich, Greg and Ranzinger, Mike and Yin, Hongxu and Lu, Yao and Kautz, Jan and Tao, Andrew and Catanzaro, Bryan and Molchanov, Pavlo},
  booktitle={Proceedings of the Computer Vision and Pattern Recognition Conference},
  pages={22487--22497},
  year={2025}
}

@inproceedings{zhai2023sigmoid,
  title={Sigmoid loss for language image pre-training},
  author={Zhai, Xiaohua and Mustafa, Basil and Kolesnikov, Alexander and Beyer, Lucas},
  booktitle={Proceedings of the IEEE/CVF international conference on computer vision},
  pages={11975--11986},
  year={2023}
}

@inproceedings{kirillov2023segment,
  title={Segment anything},
  author={Kirillov, Alexander and Mintun, Eric and Ravi, Nikhila and Mao, Hanzi and Rolland, Chloe and Gustafson, Laura and Xiao, Tete and Whitehead, Spencer and Berg, Alexander C and Lo, Wan-Yen and others},
  booktitle={Proceedings of the IEEE/CVF international conference on computer vision},
  pages={4015--4026},
  year={2023}
}

@inproceedings{caron2021emerging,
  title={Emerging properties in self-supervised vision transformers},
  author={Caron, Mathilde and Touvron, Hugo and Misra, Ishan and J{\'e}gou, Herv{\'e} and Mairal, Julien and Bojanowski, Piotr and Joulin, Armand},
  booktitle={Proceedings of the IEEE/CVF international conference on computer vision},
  pages={9650--9660},
  year={2021}
}

@inproceedings{radford2021learning,
  title={Learning transferable visual models from natural language supervision},
  author={Radford, Alec and Kim, Jong Wook and Hallacy, Chris and Ramesh, Aditya and Goh, Gabriel and Agarwal, Sandhini and Sastry, Girish and Askell, Amanda and Mishkin, Pamela and Clark, Jack and others},
  booktitle={International conference on machine learning},
  pages={8748--8763},
  year={2021},
  organization={PmLR}
}

@article{reijgwart2023efficient,
  title={Efficient volumetric mapping of multi-scale environments using wavelet-based compression},
  author={Reijgwart, Victor and Cadena, Cesar and Siegwart, Roland and Ott, Lionel},
  journal={Proceedings of Robotics: Science and System XIX},
  pages={065},
  year={2023},
  publisher={Robotics Science \& Systems Foundation}
}

@article{miller1995wordnet,
  title={WordNet: a lexical database for English},
  author={Miller, George A},
  journal={Communications of the ACM},
  volume={38},
  number={11},
  pages={39--41},
  year={1995},
  publisher={ACM New York, NY, USA}
}

@inproceedings{habitat19iccv,
  title     =     {Habitat: {A} {P}latform for {E}mbodied {AI} {R}esearch},
  author    =     {{Manolis Savva*} and {Abhishek Kadian*} and {Oleksandr Maksymets*} and Yili Zhao and Erik Wijmans and Bhavana Jain and Julian Straub and Jia Liu and Vladlen Koltun and Jitendra Malik and Devi Parikh and Dhruv Batra},
  booktitle =     {Proceedings of the IEEE/CVF International Conference on Computer Vision (ICCV)},
  year      =     {2019}
}

@inproceedings{sun2025openfrontier,
  title={OpenFrontier: General Navigation with Visual-Language Grounded Frontiers},
  author={Sun, Boyang and Cadena, Cesar and Pollefeys, Marc and Blum, Hermann},
  booktitle={IROS 2025 Workshop: Open World Navigation in Human-centric Environments},
  year      =     {2025}
}

@inproceedings{yang20253d,
  title={3D-mem: 3D scene memory for embodied exploration and reasoning},
  author={Yang, Yuncong and Yang, Han and Zhou, Jiachen and Chen, Peihao and Zhang, Hongxin and Du, Yilun and Gan, Chuang},
  booktitle={Proceedings of the Computer Vision and Pattern Recognition Conference},
  pages={17294--17303},
  year={2025}
}

@inproceedings{zhou2023navgpt,
  title={NavGPT: Explicit reasoning in vision-and-language navigation with large language models},
  author={Zhou, Gengze and Hong, Yicong and Wu, Qi},
  booktitle={Proceedings of the AAAI Conference on Artificial Intelligence},
  volume={38},
  number={7},
  pages={7359--7367},
  year={2024}
}

@article{dorbala2023can,
  title={Can an embodied agent find your “cat-shaped mug”? llm-based zero-shot object navigation},
  author={Dorbala, Vishnu Sashank and Mullen, James F and Manocha, Dinesh},
  journal={IEEE Robotics and Automation Letters},
  volume={9},
  number={5},
  pages={4083--4090},
  year={2023},
  publisher={IEEE}
}

@inproceedings{rotondi2025fungraph,
  title={Fungraph: Functionality aware 3d scene graphs for language-prompted scene interaction},
  author={Rotondi, Dennis and Scaparro, Fabio and Blum, Hermann and Arras, Kai O},
  booktitle={2025 IEEE/RSJ International Conference on Intelligent Robots and Systems (IROS)},
  pages={4083--4090},
  year={2025},
  organization={IEEE}
}
\end{document}